\documentclass[letterpaper]{article} 
\usepackage{aaai2026}  
\usepackage{times}  
\usepackage{helvet}  
\usepackage{courier}  
\usepackage[hyphens]{url}  
\usepackage{graphicx} 
\usepackage{natbib}  
\usepackage{caption} 
\usepackage{algorithm}
\usepackage{algorithmic}

\usepackage{newfloat}
\usepackage{listings}
\DeclareCaptionStyle{ruled}{labelfont=normalfont,labelsep=colon,strut=off} 
\floatstyle{ruled}
\newfloat{listing}{tb}{lst}{}
\floatname{listing}{Listing}
\title{VALA: Learning Latent Anchors for Training-Free and Temporally Consistent Video Editing}
\author{
    Zhangkai Wu\textsuperscript{\rm 1},
    Xuhui Fan\textsuperscript{\rm 1},
    Zhongyuan Xie\textsuperscript{\rm 1},
    Kaize Shi\textsuperscript{\rm 2},
    Longbing Cao\textsuperscript{\rm 1}
}
\affiliations{
    \textsuperscript{\rm 1} Macquarie University, Australia\\
    \textsuperscript{\rm 2} University of Southern Queensland, Australia
}

\usepackage{amsmath}
\usepackage{amssymb}
\newcommand{\mathbold}[1]{\ensuremath{\boldsymbol{\mathbf{#1}}}}

\newcommand{\nestedmathbold}[1]{{\mathbold{#1}}}

\newcommand{\mbc}{\nestedmathbold{c}}

\newcommand{\mbp}{\nestedmathbold{p}}

\newcommand{\mbz}{\nestedmathbold{z}}

\newcommand{\mbC}{\nestedmathbold{C}}
\newcommand{\mbD}{\nestedmathbold{D}}

\newcommand{\mbI}{\nestedmathbold{I}}

\newcommand{\mbK}{\nestedmathbold{K}}

\newcommand{\mbP}{\nestedmathbold{P}}
\newcommand{\mbQ}{\nestedmathbold{Q}}
\newcommand{\mbR}{\nestedmathbold{R}}

\newcommand{\mbV}{\nestedmathbold{V}}

\newcommand{\mbX}{\nestedmathbold{X}}

\newcommand{\mbZ}{\nestedmathbold{Z}}

\usepackage{booktabs}
\usepackage{cleveref}
\usepackage{xparse}

\begin{document}

\maketitle
\begin{abstract}
Recent advances in training-free video editing have enabled lightweight and precise cross-frame generation by leveraging pre-trained text-to-image diffusion models. However, existing methods often rely on heuristic frame selection to maintain temporal consistency during DDIM inversion, which introduces manual bias and reduces the scalability of end-to-end inference. In this paper, we propose~\textbf{VALA} (\textbf{V}ariational \textbf{A}lignment for \textbf{L}atent \textbf{A}nchors), a variational alignment module that adaptively selects key frames and compresses their latent features into semantic anchors for consistent video editing. To learn meaningful assignments, VALA propose a variational framework with a contrastive learning objective. Therefore, it can transform cross-frame latent representations into compressed latent anchors that preserve both content and temporal coherence. Our method can be fully integrated into training-free text-to-image based video editing models. Extensive experiments on real-world video editing benchmarks show that VALA achieves state-of-the-art performance in inversion fidelity, editing quality, and temporal consistency, while offering improved efficiency over prior methods.

\end{abstract}
\section{Introduction}

Recent progress in diffusion models (DM)~\cite{song2020denoising,rombach2022high,hertz2022prompt,zhang2023adding} has significantly advanced text-to-image~(T2I) generation, enabling the synthesis of high-quality, semantically coherent frame set from natural language prompts or reference image level inputs. Following the success of open source T2I models~\cite{rombach2022high}, T2I based training-free video editing~(VE) models has emerged as a promising direction. These approaches leverage pre-trained diffusion models, such as Stable Diffusion~(SD)~\cite{rombach2022high}, to perform text-driven VE without requiring additional training modules.

By introducing temporal consistency into frame-wise editing pipelines, \cite{qi_fatezero_2023,geyer2023tokenflow} can preserve the generation fidelity and cross-frame consistency by exploiting attention maps in the original U-Net architecture. Furthermore, recent training-free VE methods~\cite{li_vidtome_nodate,yang_videograin_2025} have expanded through the integration of T2I plugins, enabling more diverse and controllable edits. Compared to those fine-tuning based methods~\cite{wu2023tune,zhang_camel_2024,feng_ccedit_2024}, training-free frameworks are more efficient and broadly applicable, making them suitable for a wide range of video tasks with low computational cost.

\begin{figure}[t]
  \centering
  \includegraphics[width=0.95\columnwidth]{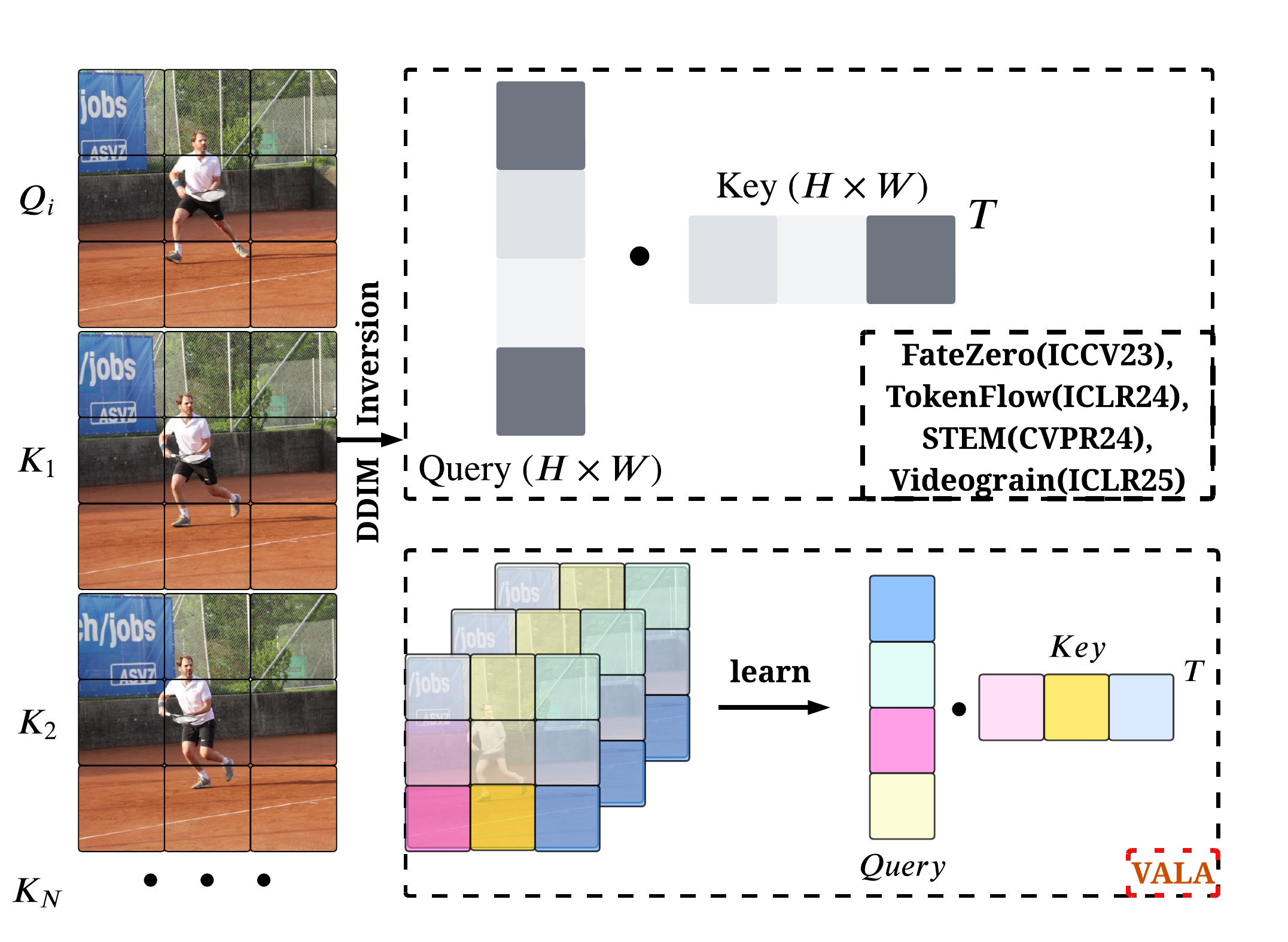}
  \caption{Comparison between existing heuristic latent frame selection methods and our variational alignment approach. Upper: heuristic methods fix reference frames without learning. Bottom: our method VALA adaptively extracts semantic frame latents via variational alignment to have compact latent anchors.}
  \label{fig:moti}
\end{figure}

A typical training-free T2I-based VE pipeline consists of two main stages: an inversion stage that maps real video frames with a reference prompt into the noise space to acquire the editing-preserved noise latents, and an editing stage that denoises the latent representations conditioned on a target prompt to generate frames aligned with user intent. To ensure temporal consistency, early methods such as FateZero~\cite{qi_fatezero_2023} and TokenFlow~\cite{geyer2023tokenflow} flatten the attention mechanism to jointly process multiple frames, thereby capturing motion continuity along the time axis in the inversion stage. 

Although it mostly ensures cross-frame consistency, \textit{how to select appropriate latent frames during inversion} still remains an open challenge. Recent methods seek to improve temporal consistency and reduce redundancy by selecting representative frames, i.e., $\{Q, K_1, K_2,...,K_N\}$ as illustrated in~\Cref{fig:moti} as latent anchors. Most approaches rely on either fixed sampling strategies~\cite{geyer2023tokenflow,qi_fatezero_2023}(e.g., the middle frame) or clustering heuristics~\cite{li_video_2024}~(e.g., K-means) to determine key latents. While these techniques offer some computational efficiency, they suffer from two key limitations. First, they lack adaptability to semantic or motion variations within the video, leading to suboptimal performance in complex scenes with abrupt transitions. Second, using manually defined static frame selection rules limits the flexibility and generalization of the learned representations.

In this paper, we present a new framework for training-free VE based on deep variational learning. Our approach introduces a principled latent alignment module, named \textit{Variational Alignment for Latent Anchors}~(VALA), which selects the key latent frame within a unified variational inference formulation. Specifically, we reformulate the alignment problem as a probabilistic latent anchor compression one. Given the latent frame of a video segment, VALA learns to assign each element to one of several latent anchor prototypes that act as compact semantic summaries across frames. Instead of reconstructing a fixed latent layout, the selected anchors are optimized under a contrastive objective to preserve both spatial fidelity and temporal coherence, providing more discriminative representations for downstream cross-frame attention modules.

Rather than relying on fixed anchor layouts or manually selected frame indices, VALA infers the most informative latent components through a learned variational posterior, enabling the dynamic set of anchors customized to each individual editing instance. This enables the model to capture video specific dynamics while accounting for variations in content, motion, and prompt semantics. The resulting anchor-based layout can be seamlessly incorporated into existing attention mechanisms, replacing dense spatial-temporal latents with a more compact and semantically structured representation.

On real-world VE benchmarks, VALA consistently outperforms existing training-free methods in terms of reconstruction fidelity, temporal consistency, and perceptual quality, and also reduces memory consumption and inference time. It achieves strong results across standard metrics, including PSNR, SSIM, CLIP similarity, and temporal warping error. User studies further indicate a clear preference for outputs generated by VALA. The ablation study highlights the selection of priors, anchor compression ratio, and assignment strategy in improving alignment and editing quality.

The main contributions can be summarized in:
\begin{itemize}
    \item \textbf{VALA}, a lightweight deep variational learning and inference framework that unifies key frame selection under a probabilistic formulation.
    \item \textbf{State of the art performance}, achieving superior reconstruction fidelity, temporal consistency, and perceptual quality compared to prior training-free VE approaches.
    \item \textbf{Uncertainty and efficiency}, offering a structured latent anchor probabilistic framework integrated with variational inference that improves temporal reasoning and reduces memory and runtime costs.
\end{itemize}

\section{Related Work}
\textbf{T2I-Based VE Models}~Due to the high inference cost of text-to-video~(T2V) models~\cite{ho2022video,singer2022make,wu2023tune} and the limited availability of high-quality video–text pairs, many recent works build on open source T2I models~\cite{rombach2022high} to enable VE guided by text or visual cues provided by users. 
Tune-A-Video~(TAV)~\cite{wu_tune--video_2023} finetunes a LoRA module that embeds temporal features to support multiframe inputs, improving controllability and editing quality. Leveraging the scalability of pre-trained LoRA modules, TAV-based approaches have advanced the integration of visual components into the base model. For example, CCEdit~\cite{feng_ccedit_2024} introduces two separate injection networks to control appearance and structure, CAMEL~\cite{zhang_camel_2024} employs CLIP-based prompt embeddings to improve motion consistency across frames, and DeCo~\cite{leonardis_deco_2025} incorporates a NeRF-based module to improve 3D perception of human subjects.
\subsubsection{T2I-Based training-free VE Models}~While TAV-based methods add trainable components, training-free approaches offer faster inference and lower resource requirements. FateZero~\cite{qi_fatezero_2023} introduces cross-frame attention modules to extract spatiotemporal attention maps for guiding multi-frame generation under varying prompts. Other methods, such as STEM~\cite{li_video_2024}, WAVE~\cite{leonardis_wave_2025}, SliceEdit~\cite{cohen_slicedit_nodate}, and COVE~\cite{wang_cove_nodate}, refine the inversion stage by selecting key frames to maintain temporal consistency. TokenFlow~\cite{geyer2023tokenflow} and Vid2Me~\cite{li_vidtome_nodate} focus on improving spatio-temporal attention maps to enhance prompt consistency during generation. Another line of work integrates external T2I plugins for fine-grained image to video control. For example, Ground-A-Video~\cite{jeong_ground--video_2024}, RAVE~\cite{kara_rave_2024}, and VideoShop~\cite{fan_vs_nodate} adopt ControlNet variants, while OCD~\cite{leonardis_object-centric_2025} and VideoGrain~\cite{yang_videograin_2025} use the segment anything model to enable object-aware VE. 

\subsubsection{Deep Variational Learning}~Variational inference in deep manner~\cite{kingma2013auto,ho2020denoising} combines stochastic variational inference~\cite{hoffman2013stochastic} with the universal approximation theorem of deep neural networks. By replacing the complex joint distribution with learned posterior parameterized by neural networks, this approach preserves uncertainty while capturing the hierarchical structure~\cite{sonderby2016ladder,van2017neural,chen2025softvq} of high-dimensional data. Rooted in Bayesian principles, variational methods approximate intractable distribution by optimizing simpler variational posterior distributions, providing a scalable alternative to traditional Markov chain Monte Carlo methods while offering more interpretable latent representations and clearer insights into model uncertainty. Such uncertainty-aware and structurally interpretable representations are largely missing in current VE methods, which often rely on fixed heuristics without probabilistic reasoning.

\section{Preliminary}

\subsubsection{Latent Encoding.} Input frames $\mbX^n \in \mathbb{R}^{N \times C \times H \times W}$ are encoded into a latent representation $\mbz^n_0 \in \mathbb{R}^{N \times c \times h \times w}$ using a pre-trained VAE encoder, where $N$ is the number of frames, $C$ is the number of input channels, and $H \times W$ denotes the original spatial resolution. The latent representation reduces the spatial size to $h \times w$ while increasing semantic capacity via a higher latent channel dimension $c$, forming the basis for subsequent inversion and editing.

\subsubsection{Prompt-Guided DDIM Inversion.} 
In pre-trained LDM, the generation proceeds by denoising step by step, predicting the clean latent $\mbz_0$ from a noisy latent $\mbz_t$. The reverse transition is parameterized as:
\begin{align*}
p_\theta(\mbz_{t-1}|\mbz_t) = \mathcal{N}\!\left(\mbz_{t-1}; \mu_\theta(\mbz_t,t), \sigma_t^2\mbI\right),
\end{align*}
where $\mu_\theta(\mbz_t,t)$ is predicted by a U-Net denoiser, and $\sigma_t^2$ is the variance schedule. For computational convenience, a noise predictor $\epsilon_\theta(\mbz_t,t)$ is used to represent the residual noise, leading to the sampling step with the target edition prompt $\mbp_{\text{tar}}$:
\begin{align*}
&\boldsymbol{z}_{t-1} \\ =&\sqrt{\frac{\alpha_{t-1}}{\alpha_t}} \boldsymbol{z}_t+\left(\sqrt{\frac{1}{\alpha_{t-1}}-1}-\sqrt{\frac{1}{\alpha_t}-1}\right) \boldsymbol{\epsilon}_\theta\left(\boldsymbol{z}_t, t,\phi(\mbp_{\text{tar}})\right),
\end{align*}
where $\alpha_t$ is the diffusion coefficient and $\phi(\cdot)$ is the text encoder.

To enable semantic editing without discarding structural content, a multi-step DDIM inversion~\cite{song2020denoising,nichol2021improved} is adopted to deterministically map the clean latent $\mbz^n_0$ to a semantically aligned noisy latent $\mbz^n_t$. With the source prompt $\mbP_{\texttt{src}}$, each inversion step can be written as:
\begin{align*}
\boldsymbol{z}_{t}&=
\sqrt{\frac{\alpha_{t}}{\alpha_t-1}} \boldsymbol{z}_{t-1}\\+&\left(\sqrt{\frac{1}{\alpha_{t}}-1}-\sqrt{\frac{1}{\alpha_{t-1}}-1}\right) \boldsymbol{\epsilon}_\theta\left(\boldsymbol{z}_{t-1}, t-1, \phi(\mbp_{\text{src}})\right).
\end{align*}
For notational brevity, the frame index superscript $n$ is omitted in the remainder of this subsection.


To ensure the inversion latents aligned with source prompt $\mbp_{\text{src}}$, classifier-free guidance with null-text technique~\cite{ho2020denoising,mokady2023null} is employed to interpolates between conditional and unconditional predictions to guarantee inversion fidelity with text condition:
\begin{multline}
\tilde{\epsilon}_\theta(\mbz_{t-1},t-1) =w\,\epsilon_\theta(\mbz_{t-1},t-1,\phi(\mbp_{\text{src}}))\\  
+ (1-w)\,\epsilon_\theta(\mbz_{t-1},t-1,\phi(\varnothing))\nonumber
\end{multline}
where $w$ controls semantic strength. The resulting latent $\mbz_T$ preserves the spatial details of the original frames while aligning with the source prompt $\mbp_{\text{src}}$. 


\subsubsection{Modeling Temporal Coherence via Cross-Frame Self-Attention.} Temporal consistency plays a central role in coherent VE. Previous approaches introduce hand-crafted spatial-temporal alignment functions $\varphi(\cdot)$ to reorganize latent features from neighboring frames into attention-compatible formats~\cite{qi_fatezero_2023,geyer2023tokenflow,li_video_2024}. These mappings are typically designed to capture inter-frame continuity by reshaping the input into structured latents.

Given latents $\mbz_t \in \mathbb{R}^{\ell \times c \times h \times w}$ from $\ell$ selected frames, the alignment function permutes and reshapes as:
\begin{equation}
\varphi: \mathbb{R}^{\ell \times c \times h \times w} \rightarrow \mathbb{R}^{(h \cdot w) \times \ell \times c},
\label{eq:allign}
\end{equation}
allowing self-attention to operate across frames at each spatial location. To compute the cross-frame self-attention matrix, the transformed latents are projected into query, key, and value spaces:
\begin{equation}
\mbQ = W_\mbQ \cdot \varphi(\mbz_t), \quad
\mbK = W_\mbK \cdot \varphi(\mbz_t), \quad
\mbV = W_\mbV \cdot \varphi(\mbz_t),
\end{equation}
where $W_\mbQ, W_\mbK, W_\mbV \in \mathbb{R}^{c \times d}$ are learnable parameters. This mechanism enables each spatial location to aggregate temporal information from aligned positions across frames, promoting structural and motion continuity.

However, handcrafted operators $\varphi(\cdot)$ often struggle to capture complex motion, especially when such motion involves delays or inconsistent dynamics. Moreover, incorporating too many frames can lead to latent representation redundancy, reducing the effectiveness of self-attention in modeling meaningful temporal dependencies.

\begin{figure*}[t]
    \centering
    \includegraphics[width=2\columnwidth]{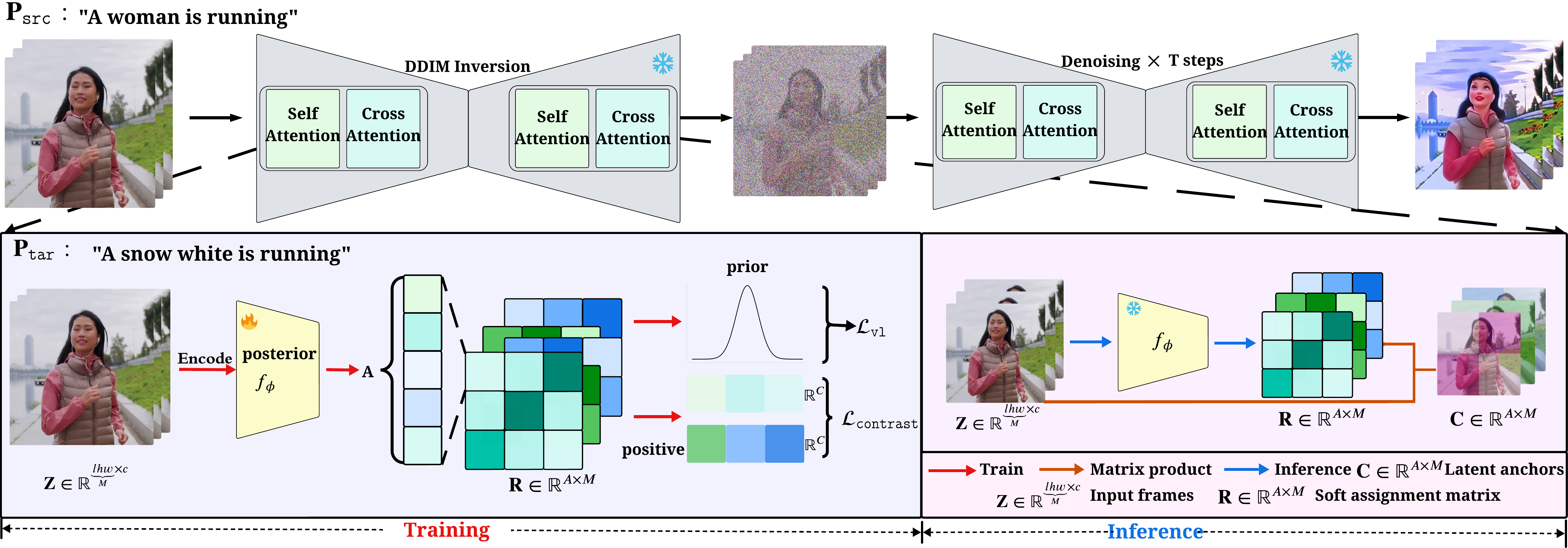}  
    \caption{The training and inference process of VALA. Instead of relying on fixed alignment functions $\varphi(\cdot)$ in~\Cref{eq:allign}, VALA selects a compact and meaningful latent anchors to represent the spatiotemporal structure of a video learned by our posterior $f_{\phi}$. In the training stage, a lightweight network parameterized by $\phi$ is optimized by automatically learn the soft assignment matrix $\mbR$. In the inference stage, the latent anchors $\mbC$ is assigned by learned from soft assignment matrix and input latent representation $\mbZ$ to support the following edition process. }
    \label{fig:fram}
\end{figure*}

\section{Latent Compression via Attention Weighting Network}
\label{sec:vala-weight}

Given a latent tensor $\mbz_t \in \mathbb{R}^{\ell \times c \times h \times w}$ from $\ell$ input frames, we first flatten it into a input latent representation matrix $\mathbf{Z} \in \mathbb{R}^{M \times c}$, where $M = \ell \cdot h \cdot w$ and $c$ is the feature dimension.

We introduce a lightweight neural network $f_\phi(\cdot)$ with DCGAN architecture encoders~\cite{radford2015unsupervised,burgess2018understanding,shao2020controlvae} that maps each latents $\mathbf{z}_m \in \mathbb{R}^c$ to a assignment attention layout matrix over $A$ anchors:
\begin{equation}
r_{am} = \mathrm{softmax}(f_\phi(\mathbf{z}_m))_a, \quad \text{with} \quad \sum_{a=1}^{A} r_{am} = 1,
\end{equation}
where attention matrix $\mathbf{R} \in \mathbb{R}^{A \times M}$ denotes the resulting soft assignment matrix. To summarize the full latent sequence into $A$ semantic anchors, we apply matrix multiplication to get latent anchors $\mbC$:
\begin{equation}
\mathbf{C} = \mathbf{R} \cdot \mathbf{Z} \in \mathbb{R}^{A \times c},
\end{equation}
where each row $\mathbf{c}_a$ captures a compressed semantic representation $\mathbb{R}^{c}$ learned from the input.

\subsubsection{Variational Regularization.}  
To prevent the model from producing degenerate assignments, such as collapsing all latent features onto a few anchors, we introduce a regularization term based on KL divergence $\mbD_{\mathrm{KL}}$. This term encourages the learned assignment distribution to stay close to a prior distribution, thereby improving generalization and anchor usage diversity.

We treat each assignment vector $\mathbf{r}_m \in \mathbb{R}^{A}$ as a categorical distribution over $A$ anchors for a given latent feature $\mathbf{z}_m$. This vector is obtained through a softmax transformation from the output of the assignment network $f_{\phi}$. We define the posterior as:
\begin{equation}
q(a \mid \mathbf{z}_m) := \mathbf{r}_m = \mathrm{softmax}(f_{\phi}(\mathbf{z}_m)).
\end{equation}
and we set the prior distribution $p(a)$ to be uniform over the $A$ anchors:
\begin{equation}
p(a) = \frac{1}{A}, \quad \text{for all } a = 1, \dots, A.
\end{equation}
The KL divergence between the posterior $q(a \mid \mathbf{z}_m)$ and the uniform prior $p(a)$ is given by:
\begin{equation}
\mathcal{L}_{\mathrm{VI}}^{(m)} = \mbD_\mathrm{KL}(q(a \mid \mathbf{z}_m) \parallel p(a)) = \sum_{a=1}^{A} r_{am} \cdot \log \left( \frac{r_{am}}{1/A} \right).
\end{equation}
The analytical derivation of the $\mbD_{\mathrm{KL}}$ is given as follows:
\begin{equation}
\mathcal{L}_{\mathrm{VI}}^{(m)} = \sum_{a=1}^{A} r_{am} \cdot \log \left( r_{am} \cdot A \right).
\end{equation}

This loss penalizes sharp or overly concentrated assignments, and encourages the attention weights $\mathbf{R}$ to cover all anchors more evenly during training. By doing so, the model avoids biasing towards a small subset of anchors and ensures that each anchor has a chance to participate in representing the input. Since $A \ll M$, this uniform coverage does not hindering us to obtain a compressed latent anchors.

\subsubsection{Contrastive Loss.}  
To make the compressed token $\mathbf{c}_a$ semantically aligned with its most meaningful latents, we identify the top-$k$ input features with highest $r_{am}$ and encourage their similarity:
\begin{equation}
\mathcal{L}_{\text{contrast}}^{(a)} = \frac{1}{k} \sum_{m \in \mathcal{P}_a} -\log \frac{\exp(\text{sim}(\mathbf{c}_a, \mathbf{z}_m) / \tau)}{\sum_{m'=1}^{M} \exp(\text{sim}(\mathbf{c}_a, \mathbf{z}_{m'}) / \tau)},
\end{equation}
where $\mathcal{P}_a$ is the set of top-$k$ tokens assigned to anchor $a$, and $\text{sim}(\cdot,\cdot)$ denotes cosine similarity.

\subsubsection{Training Objective of VALA.}~The final training objective combines the contrastive and regularization terms:
\begin{equation}
\mathcal{L}_{\text{VALA}} = \sum_{a=1}^{A} \mathcal{L}_{\text{contrast}}^{(a)} + \lambda_{\text{VI}} \cdot \mathcal{L}_{\text{VI}}.
\label{eq:loss}
\end{equation}

\subsubsection{Inference of VALA.}~During inference, the learned attention network $f_\phi$ is applied to a new input sequence $\mathbf{z}_t \in \mathbb{R}^{\ell \times c \times h \times w}$. We first flatten $\mathbf{z}_t$ along the spatial and temporal dimensions into a matrix $\mathbf{Z} \in \mathbb{R}^{M \times c}$, where $M = \ell \cdot h \cdot w$. Then, for each latent representation $\mathbf{z}_m$, we compute a soft assignment over $A$ anchors using the trained model $f_{\phi}$:
\begin{equation}
r_{am} = \mathrm{softmax}(f_\phi(\mathbf{z}_m))_a.
\end{equation}

All assignments are collected into the attention assignment weight matrix $\mathbf{R} \in \mathbb{R}^{A \times M}$. The final compressed latent anchors is obtained via a single matrix multiplication:
\begin{equation}
\mathbf{C} = \mathbf{R} \cdot \mathbf{Z} \in \mathbb{R}^{A \times c}.
\end{equation}

Each row $\mathbf{c}_a$ in $\mathbf{C}$ serves as a semantic anchor that summarizes relevant anchors throughout the video segment. Since both the attention assignment matrix and latent matrix are computed once, the process is efficient and free of iterative refinement. The result $\mathbf{C}$ can be directly used in downstream modules as a compact and structured representation for semantic editing.

Compared with using the full latent $\mathbf{Z}$, this inference pipeline reduces memory consumption and improves uncertainty by focusing on a small number of learned anchors. The time efficiency and memory usage are systematically evaluated in the following sections and the whole process can be illustrated in~\Cref{fig:fram}.

\section{Experiments}

\subsection{Experimental Setup}

We evaluated our variational VE framework on the DAVIS dataset~\cite{pont20172017} and a set of various Internet videos with varying content and length. Videos are $512 \times 512, 640 \times 320$ resolution before processing. We adopt SD 1.5~\cite{rombach2022high} as the generative backbone, and apply DDIM inversion~\cite{song2020denoising} with 50 steps and a classifier-free guidance scale of 7.5. All experiments are conducted on a single NVIDIA H100 GPU. The variational encoder $f_\phi$ is optimized using the Adam optimizer with a learning rate of $1 \times 10^{-4}$.  

\subsection{Metrics and Baselines}

To evaluate the effectiveness of our variational inversion strategy, we compare it with several representative training-free T2I based VE baselines that differ in how they perform inversion and select key frames. Specifically, we consider TokenFlow~\cite{geyer2023tokenflow}, which relies on nearest neighbor token propagation from a manually selected reference frame; and FateZero~\cite{qi_fatezero_2023}, which selects two adjacent frames and fuses their spatial-temporal attention without explicit motion estimation and STEM~\cite{li_video_2024}, which selects representative tokens via a fixed-basis EM algorithm to guide inversion. 

For quantitative evaluation, we assess both the fidelity of inverted frames and the consistency and quality of edited outputs. Reconstruction quality is measured using PSNR and SSIM, which reflect pixel-level accuracy and perceptual similarity, respectively. Editing quality is evaluated through the CLIP Score, which captures alignment between edited frames and the text prompt in the CLIP embedding space~\cite{radford2021learning}, and Warp Error~\cite{teed2020raft}, which estimates temporal consistency by comparing generated frames to optical-flow-based warped frames using RAFT~\cite{teed2020raft}. These metrics together provide a comprehensive assessment of how well each method preserves content and structure during inversion and how effectively it enables faithful and consistent VE.

\subsection{Inversion Quality Comparison}

\begin{figure*}[!htbp]
    \centering
    \includegraphics[width=0.9\linewidth]{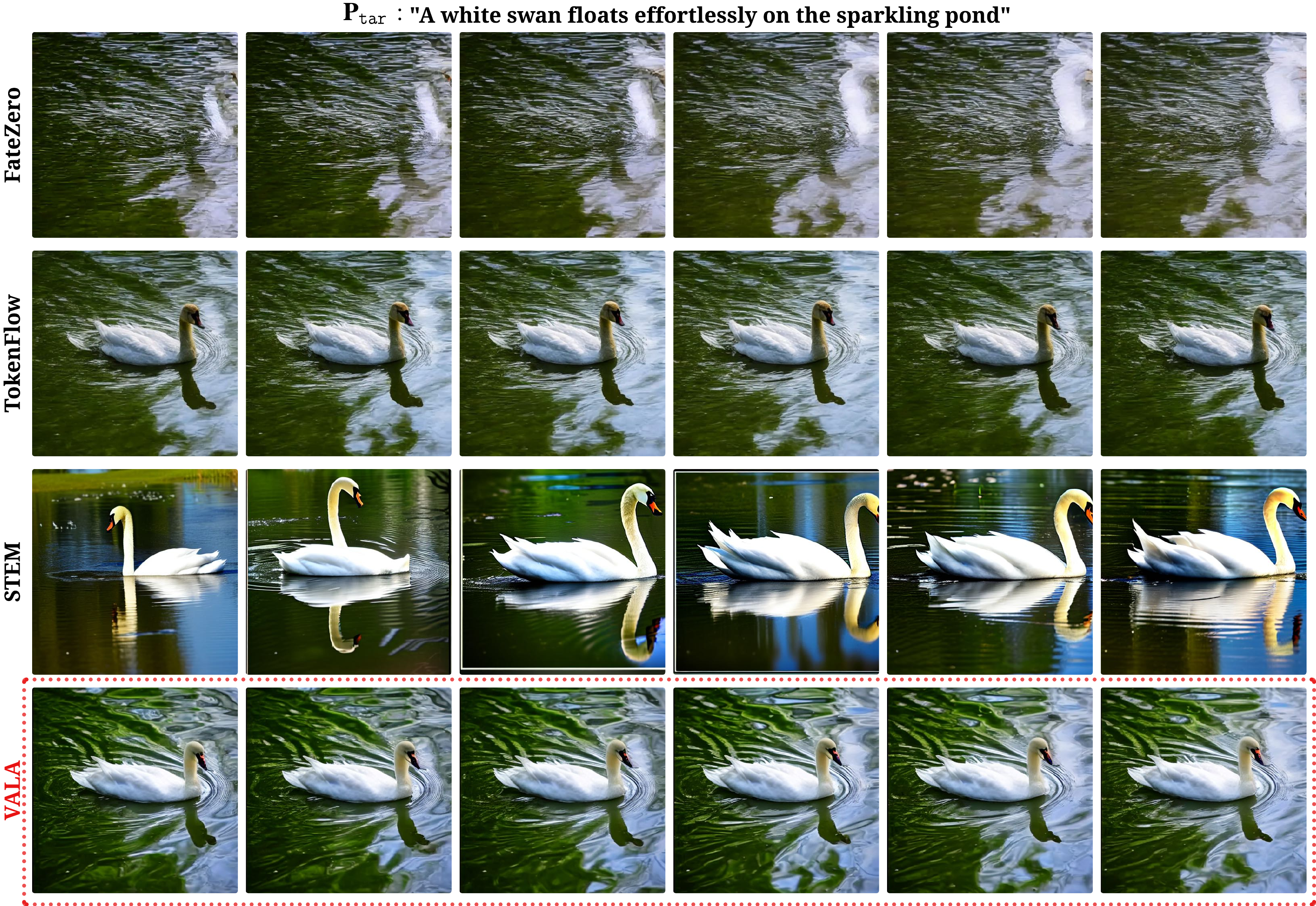}
    \caption{Qualitative comparison of editing results. Our variational inversion preserves temporal coherence and improves object fidelity under challenging prompts.}
    \label{fig:edit_vis}
\end{figure*}

We compare our method with existing video inversion baselines, focusing on reconstruction fidelity and computational cost. As shown in~\Cref{tab:inv_quality}, firstly, our variational inversion achieves the best quantitative performance. It attains the highest PSNR (26.18) and SSIM (0.875), indicating superior fidelity in both pixel-wise accuracy and structural similarity. Compared to STEM, which relies on a fixed anchor set with EM assignment, our adaptive anchor mechanism leads to slightly better reconstruction using fewer tokens.

Finally, in terms of runtime, our method is competitive with STEM while significantly outperforming TokenFlow and FateZero. Although TokenFlow and FateZero appear faster in absolute time, they scale poorly with sequence length because they rely on full-resolution attention or dense pixel alignment. In contrast, our model learns to compress latent representations into a compact set of latent anchors ($A \ll M$), enabling fast inversion and reconstruction. Unlike STEM, our method avoids iterative EM steps, which further improves time efficiency.

\begin{table}[ht]
\centering
\begin{tabular}{l|c|c|c}
\toprule
Method & PSNR (↑) & SSIM (↑) & Time (min) \\
\midrule
TokenFlow (DDIM) & 23.52 & 0.810 & 0.82 \\
FateZero (2-frame) & 24.21 & 0.834 & 1.12 \\
STEM (K=256, EM) & 25.94 & 0.869 & 1.50 \\
    \midrule
Ours (VI, adaptive) & \textbf{26.18} & \textbf{0.875} & \textbf{1.46} \\
\bottomrule
\end{tabular}
\caption{Comparison of video inversion methods.}
\label{tab:inv_quality}
\end{table}

\subsection{Editing Performance}

We integrate our variational inversion into existing editing pipelines, replacing the default DDIM inversion used by TokenFlow and FateZero. Quantitative results are shown in ~\Cref{tab:edit_perf}, where our method consistently improves both edit fidelity (CLIP score) and temporal consistency (Warp Error). In particular, Ours + TokenFlow achieves a CLIP score of 0.33 and a warp error of 3.8, outperforming its original counterpart (0.31 / 4.9). Similarly, our method enhances FateZero from 0.29 / 7.2 to 0.31 / 4.6, showing improved edit reliability and temporal smoothness.

These improvements highlight the benefit of variational token alignment in preserving semantic structure across frames. Unlike original inversion schemes that treat each frame independently, our compact alignment-aware layout promotes consistency.

We also visualize the editing results in~\Cref{fig:edit_vis}, where our inversion process better preserves object shape and motion coherence, particularly in challenging scenes with complex prompts, e.g.,\texttt{"floats effortlessly"} in $\mbp_{\mathrm{tar}}$. Compared to traditional methods such as FateZero and TokenFlow (rows 1 and 2 in~\Cref{fig:edit_vis}), the generated results exhibit apparent corruption and fail to align well with the intended prompts. In contrast to STEM (row 3 in~\Cref{fig:edit_vis}), our method maintains better temporal consistency, as the first frame in STEM shows a mismatched direction and inconsistencies along the time axis.

\begin{table}[ht]
\centering

\begin{tabular}{l|c|c}
\toprule
Method & CLIP Score (↑) & Warp Error (↓) \\
\midrule
TokenFlow & 0.31 & 4.9 \\
Ours + TokenFlow & \textbf{0.33} & \textbf{3.8} \\
FateZero & 0.29 & 7.2 \\
Ours + FateZero & \textbf{0.31} & \textbf{4.6} \\
\bottomrule
\end{tabular}

\caption{Zero-shot video editing performance.}
\label{tab:edit_perf}
\end{table}

\begin{figure*}[!htbp]
\centering
\includegraphics[width=0.9\linewidth]{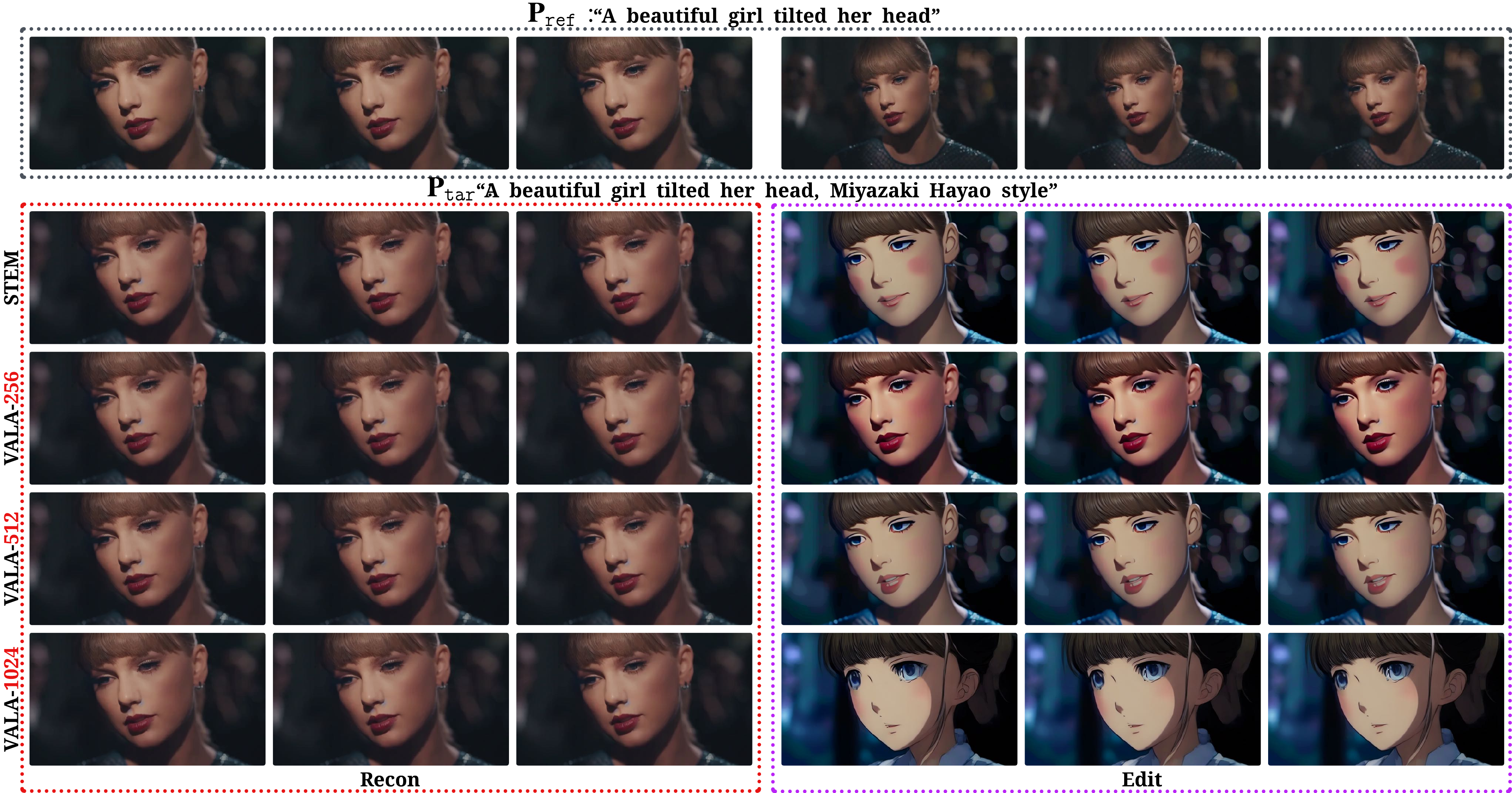}
\caption{Effect of the anchor number $A$ on inversion and editing performance. Performance improves with $A$, but corrupt beyond $A=512$. Left part is the reconstruction and right part is editing.}
\label{fig:ablation_a}
\end{figure*}

\subsection{User Study}

Following the protocols in previous methods~\cite{li_video_2024,yang_videograin_2025}, to further validate the perceptual quality of our method, we conduct a user study comparing our variational inversion with the default DDIM-based inversion used in TokenFlow and FateZero. We recruited 20 participants and randomly selected 12 edited video pairs per participant for evaluation. For each pair, participants were asked to choose: (a) the version with better editing fidelity, and (b) the one with more consistent temporal motion.

As shown in ~\Cref{tab:user_study}, our inversion consistently receives higher preference across both aspects. In particular, users preferred our inversion over TokenFlow in 75\% of cases for fidelity and 68\% for consistency. Similarly, when paired with FateZero, our method was favored in 78\% and 70\% of cases, respectively.

\begin{table}[ht]
\centering

\begin{tabular}{l|c|c}
\toprule
Comparison & Better Fidelity & Better Consistency \\
\midrule
Ours vs TokenFlow & \textbf{75\%} & \textbf{68\%} \\
Ours vs FateZero & \textbf{78\%} & \textbf{70\%} \\
\bottomrule
\end{tabular}

\caption{User preference results comparing variational inversion and DDIM-based inversion.}
\label{tab:user_study}
\end{table}

\subsection{Effectiveness of Variational Priors}
\label{subsec:dis}
~To examine how variational priors contribute to our method’s performance, we conduct an ablation study comparing three settings: (1)~no prior, (2)~only gaussian prior, and (3)~categorical priors. Our goal is to assess whether priors improve anchor assignment quality, anchor space regularity, and downstream inversion/editing performance. We report quantitative results on reconstruction and editing metrics in ~\Cref{tab:prior_ablation}.

We find that the Gaussian prior helps encourage balanced usage of anchor tokens, leading to better anchor grouping and reduced posterior collapse~\cite{zhao2017infovae}. The Categorical prior further improves the consistency and separability of anchor embeddings by regularizing the latent space and lead to improvements in PSNR, SSIM, and CLIP Score, while also reducing temporal Warp Error.

\begin{table}[ht]
\centering
\begin{tabular}{l|c|c|c|c}
\toprule
Prior Setting & PSNR ↑ & SSIM ↑ & CLIP ↑ & Warp ↓ \\
\midrule
No Prior & 24.91 & 0.851 & 0.30 & 5.7 \\
Gaussian & 25.64 & 0.862 & 0.32 & 4.9 \\
Categorical & \textbf{26.18} & \textbf{0.875} & \textbf{0.33} & \textbf{3.8} \\
\bottomrule
\end{tabular}

\caption{Effect of variational priors on inversion and editing performance.}
\label{tab:prior_ablation}
\end{table}

\textbf{Computational Efficiency and Complexity Analysis} ~A key motivation behind our variational design is to reduce the computational overhead of dense attention over long video sequences. Instead of computing attention across all spatiotemporal latents, we introduce a compact set of $K$ latent anchors that serve as shared key-value representations. This design greatly improves scalability.

The standard self-attention mechanism has a time complexity of $\mathcal{O}(M^2 C)$, where $M = \ell \cdot h \cdot w$ is the number of spatiotemporal tokens, and $C$ is the channel dimension. In contrast, our method reduces this to $\mathcal{O}(M A C)$ by replacing the quadratic attention with attention between $M$ tokens and $A$ anchors, where $A \ll M$. The additional cost from the anchor selection network $f_\phi$ is linear, only $\mathcal{O}(M C)$.

To evaluate actual efficiency, we measure runtime and peak memory usage across various video lengths. As shown in ~\Cref{tab:runtime_efficiency}, our method consistently achieves lower computation time and memory usage compared to TokenFlow and FateZero. This advantage becomes more significant as video length increases.

\begin{table}[ht]
\centering

\begin{tabular}{l|c|c}
\toprule
Method & Time (min) ↓ & GPU Memory (GB) ↓ \\
\midrule
TokenFlow & 0.82 & 9.4 \\
FateZero & 1.12 & 12.0 \\
Ours (Variational) & \textbf{0.46} & \textbf{6.2} \\
\bottomrule
\end{tabular}

\caption{Runtime and peak memory comparison on $640 \times 320$ pixel videos.}
\label{tab:runtime_efficiency}
\end{table}

Overall, the proposed variational formulation not only improves the semantic quality of attention representations but also enables linear complexity with respect to video length, making it efficient and scalable for practical use.

\textbf{Ablation Study: Effect of Anchor Number $A$} ~We perform an ablation study to evaluate how the number of selected anchors $A$ impacts the reconstruction and editing performance. We test values of $A \in \{256, 512, 1024\}$ on videos with resolution $640 \times 320$ and frame length $N = 60$. The results are summarized in Figure~\ref{fig:ablation_a}.

We observe that increasing $A$ generally improves both PSNR and CLIP score, as a larger anchor set provides a finer-grained basis for representing spatial-temporal information. However, the improvement degrade around $A=1024$ which means $A=1024$ yields only marginal gains while doubling computational cost. When $A=256$, the model still reconstructs video content reasonably well, but editing quality is degraded, indicating under-representation of key content. Therefore, we adopt $A=512$ as the default configuration, offering a strong trade-off between performance and efficiency.

\section{Conclusion}

In this work, we introduced VALA, a novel T2I based training-free framework for VE that leverages variational latent alignment to address the challenge of temporal inconsistency. By reformulating keyframe selection as a latent anchor assignment problem, VALA dynamically learns compact and semantically structured latent  representations via a trained variational inference process integrated with a contrastive learning objective. This enables improved temporal coherence, better editing fidelity, and enhanced interpretability without requiring modifications to the underlying latent diffusion model. Extensive experiments demonstrate that VALA achieves SOTA performance in both inversion and editing tasks across standard benchmarks, while remaining computationally efficient. Beyond performance gains, VALA offers a new perspective on integrating deep variational learning with T2I based training-free VE models, opening new directions for future work for latent representation learning in VE, uncertainty-aware editing, and interpretable video manipulation.

\clearpage
\newpage

\bibliography{aaai2026}


\clearpage
\newpage
\section{Appendix}
\subsection{Implementation Details of the VALA Module}

The \textbf{VALA} (\textbf{V}ariational \textbf{A}lignment for \textbf{L}atent \textbf{A}nchors) module performs latent alignment and compression in two main steps: variational assignment of tokens to anchors and aggregation of anchor embeddings via soft pooling. Below, we describe the implementation details and formalize each step.

\textbf{Variational Assignment of Latent Tokens}~Given the input latent tensor $ \mbz_t \in \mathbb{R}^{\ell \times c \times h \times w} $ from $ \ell $ frames, we reshape it into a token matrix:
\begin{equation}
\mbz \in \mathbb{R}^{M \times c}, \quad \text{where } M = \ell \cdot h \cdot w.
\end{equation}

A lightweight network $ f_\phi $ estimates soft assignments between tokens and $ A $ anchor slots by producing a responsibility matrix $ \mathbf{R} \in \mathbb{R}^{A \times M} $, with each column normalized via softmax:
\begin{equation}
r_{am} = \mathrm{softmax}(f_\phi(\mbz_m))_a, \quad \text{subject to } \sum_{a=1}^{A} r_{am} = 1.
\label{eq:softmax}
\end{equation}

This step maps high-resolution spatiotemporal features into a lower-dimensional anchor space while preserving semantic alignment.

\textbf{Anchor Construction via Responsibility-Weighted Pooling}~Each anchor vector $ \hat{\mbz}_a \in \mathbb{R}^{c} $ is formed by aggregating token features according to their assignment weights:
\begin{equation}
\mbz_a = \sum_{m=1}^{M} r_{am} \cdot \mbz_m.
\end{equation}

This can be expressed in matrix form as:
\begin{equation}
\mbc_{1:A} = \mathbf{R} \cdot \mbz \in \mathbb{R}^{A \times c},
\label{eq:anchor-pool}
\end{equation}
where the resulting anchor matrix serves as a compact representation of the original sequence. These anchors are subsequently used in attention or decoding modules for temporally consistent editing.

\subsection{KL Divergence under Gaussian Prior}
\label{app:kl-gaussian}

In this section, we elaborate on the use of Gaussian priors for regularizing the latent assignment space introduced in~\Cref{subsec:dis}. In our model, the attention-based assignment mechanism produces a set of anchor-level latent embeddings $\mathbf{C}$ via a soft clustering over flattened token features $\mathbf{Z}$. To avoid degenerate clustering behavior such as mode collapse or overconfident assignments, we apply a variational regularization term that encourages the latent encoding to align with a prior distribution in the anchor space.

To formalize this, we consider a latent vector $\mathbf{c}_a \in \mathbb{R}^c$ (i.e., one of the compressed anchor embeddings) as a random variable drawn from a posterior distribution $q(\mathbf{z}) = \mathcal{N}(\boldsymbol{\mu}, \mathrm{diag}(\boldsymbol{\sigma}^2))$. We place a standard Gaussian prior $p(\mathbf{z}) = \mathcal{N}(\mathbf{0}, \mathbf{I})$ on each anchor to serve as a regularizing reference. The KL divergence between the posterior and the prior is then computed as:
\begin{equation}
\mathbb{D}_{\mathrm{KL}}(q(\mathbf{z}) \| p(\mathbf{z})) = \frac{1}{2} \sum_{i=1}^{c} \left( \sigma_i^2 + \mu_i^2 - 1 - \log \sigma_i^2 \right).
\end{equation}

This closed-form expression penalizes anchor embeddings that deviate significantly from the unit Gaussian in either their mean (via $\mu_i^2$) or variance (via $\sigma_i^2$), and prevents posterior collapse by enforcing entropy through the $-\log \sigma_i^2$ term. In our context, this term encourages each anchor to represent diverse semantic content without collapsing into a small subregion of the latent space. Combined with the attention weighting mechanism, this KL loss facilitates compact yet expressive latent compression, improving generalization and preventing trivial solutions in anchor assignments.

\begin{algorithm}[ht]
\caption{\textbf{VALA Training} — Joint Optimization of Assignment Network and Anchor Representations}
\label{alg:vala-train}
\begin{algorithmic}[1]
\REQUIRE Latent tensor $\mathbf{z}_t \in \mathbb{R}^{\ell \times c \times h \times w}$, number of anchors $A$, top-k parameter $k$, temperature $\tau$, regularization weight $\lambda_{\text{VI}}$
\ENSURE Trained assignment network $f_\phi$
\STATE \textit{Reshape latent tensor into matrix form:}
\STATE $\mathbf{Z} \gets \text{reshape}(\mathbf{z}_t) \in \mathbb{R}^{M \times c}$, where $M = \ell \cdot h \cdot w$
\STATE \textit{Compute soft assignment matrix:}
\STATE $r_{am} = \text{softmax}(f_\phi(\mathbf{z}_m))_a$ for all $m, a$
\STATE Collect into attention matrix $\mathbf{R} \in \mathbb{R}^{A \times M}$
\STATE \textit{Generate compressed anchor representations:}
\STATE $\mathbf{C} = \mathbf{R} \cdot \mathbf{Z} \in \mathbb{R}^{A \times c}$
\STATE \textit{Compute KL regularization loss:}
\STATE $\mathcal{L}_{\text{VI}} = \sum_{m=1}^{M} \sum_{a=1}^{A} r_{am} \cdot \log(r_{am} \cdot A)$
\STATE \textit{Compute contrastive loss for each anchor:}
\FOR{$a = 1$ to $A$}
    \STATE Find top-k tokens: $\mathcal{P}_a = \{\text{top-}k \text{ indices } m \text{ with highest } r_{am}\}$
    \STATE $\mathcal{L}_{\text{contrast}}^{(a)} = \frac{1}{k} \sum_{m \in \mathcal{P}_a} -\log \frac{\exp(\text{sim}(\mathbf{c}_a, \mathbf{z}_m) / \tau)}{\sum_{m'=1}^{M} \exp(\text{sim}(\mathbf{c}_a, \mathbf{z}_{m'}) / \tau)}$
\ENDFOR
\STATE \textit{Compute total loss and update parameters:}
\STATE $\mathcal{L}_{\text{VALA}} = \sum_{a=1}^{A} \mathcal{L}_{\text{contrast}}^{(a)} + \lambda_{\text{VI}} \cdot \mathcal{L}_{\text{VI}}$
\STATE Backpropagate gradients and optimize $f_\phi$
\end{algorithmic}
\end{algorithm}

\begin{algorithm}[ht]
\caption{\textbf{VALA Inference} — Extraction of Compact Anchor Representation}
\label{alg:vala-infer}
\begin{algorithmic}[1]
\REQUIRE Latent tensor $\mathbf{z}_t \in \mathbb{R}^{\ell \times c \times h \times w}$, trained assignment network $f_\phi$, number of anchors $A$
\ENSURE Compressed anchor representations $\mathbf{C} \in \mathbb{R}^{A \times c}$
\STATE \textit{Reshape latent tensor into matrix form:}
\STATE $\mathbf{Z} \gets \text{reshape}(\mathbf{z}_t) \in \mathbb{R}^{M \times c}$, where $M = \ell \cdot h \cdot w$
\STATE \textit{Compute soft assignment probabilities:}
\STATE $r_{am} = \text{softmax}(f_\phi(\mathbf{z}_m))_a$ for all $m \in \{1, \ldots, M\}, a \in \{1, \ldots, A\}$
\STATE Collect into attention matrix $\mathbf{R} \in \mathbb{R}^{A \times M}$
\STATE \textit{Generate compressed anchor representation:}
\STATE $\mathbf{C} = \mathbf{R} \cdot \mathbf{Z} \in \mathbb{R}^{A \times c}$
\RETURN $\mathbf{C}$ as compact latent representation for downstream tasks
\end{algorithmic}
\end{algorithm}

\end{document}